\documentclass{article}

\usepackage[preprint, nonatbib]{neurips_2021}

\usepackage[utf8]{inputenc} 
\usepackage[T1]{fontenc}    
\usepackage{hyperref}       
\usepackage{url}            
\usepackage{booktabs}       
\usepackage{amsfonts}       
\usepackage{nicefrac}       
\usepackage{microtype}      
\usepackage{xcolor}         

\usepackage[british]{babel}

\usepackage[numbers]{natbib} 
\usepackage{mathtools} 
\usepackage{siunitx} 
\usepackage{booktabs} 
\usepackage{tikz} 
\usepackage{todonotes}

\usepackage{bm}
\usepackage{amsmath}
\usepackage{amssymb}
\usepackage{makeidx}
\usepackage{graphicx}
\usepackage{siunitx}
\usepackage{color}
\usepackage{colortbl}
\usepackage{gensymb}
\usepackage{subcaption}
\usepackage{multirow}
\usepackage{paralist}

\usepackage{diagbox}
\usepackage{float}

\usepackage{subcaption}

\definecolor{mydarkblue}{rgb}{0,0.08,0.45}

\usepackage[ruled, linesnumbered]{algorithm2e}

\SetCommentSty{mycommfont}
	
\newcommand{\vecVar}[1]{\mathbf{#1}}
\newcommand{\matVar}[1]{\mathbf{#1}}

\newcommand{\xvec}[0]{\vecVar{x}}

\newcommand{\timeVar}[0]{t}

\newcommand{\methodUni}[0]{ssGP}
\newcommand{\methodUniRob}[0]{r-\methodUni}
\newcommand{\method}[0]{ssGPFA}
\newcommand{\methodRob}[0]{r-\method}

\newcommand{\rrcf}{RRCF}
\newcommand{\htm}{HTM}
\newcommand{\omni}{OmniAnomaly}

\newcommand{\B}[1]{\mathbf{#1}}

\newcommand{\R}{\mbox{\(\mathbb R\)}}

\newcommand{\N}{\mbox{\(\mathcal N\)}}

\newcommand{\inv}{^{-1}}

\newtheorem{definition}{Definition}

\title{Online Time Series Anomaly Detection with State Space Gaussian Processes}

\author{
	Christian~Bock\thanks{These authors contributed equally} \\
	ETH Zurich\textsuperscript{1}
	\And
	Fran\c{c}ois-Xavier~Aubet\textsuperscript{$*$} \\
	Amazon Research	
	\AND
	Jan Gasthaus \\
	Amazon Research 
	\And
	Andrey Kan \\
	Amazon Research
	\And
	Ming Chen \\
	Amazon Research
	\And 
	Laurent Callot \\
	Amazon Research
}

\begin{document}
\maketitle
\footnotetext[1]{Work completed while interning at Amazon Research}
\setcounter{footnote}{1}
\begin{abstract}

%
%
We propose \methodRob, an unsupervised online anomaly detection model for uni- and multivariate time series building on the efficient state space formulation of Gaussian processes.
%
%
For high-dimensional time series, we propose an extension of Gaussian process factor analysis to identify the common latent processes of the time series, allowing us to detect anomalies efficiently in an interpretable manner.
We  gain explainability while speeding up computations by imposing an orthogonality constraint on the mapping from the latent to the observed.
Our model's robustness is improved by using a simple heuristic to skip Kalman updates when encountering anomalous observations.
We investigate the behaviour of our model on synthetic data and show on standard benchmark datasets that our method is competitive with state-of-the-art methods while being computationally cheaper. 

\end{abstract}

\section{Introduction}\label{sec:intro}


Online anomaly detection~(AD) in time series data has a wide range of applications, enabling e.g.\ automatic monitoring \& alarming, quality control, and predictive maintenance \citep{hsieh2019unsupervised, ren2019time, gao2020robusttad, cook2019anomaly}.
One commonly used ansatz to formalise the intuitive notion of an \emph{anomaly} as something that is different from the ``normal behavior'', is to define it as an event that is improbable under a probabilistic model of the data. This approach is particularly appealing in the time series setting, where probabilistic models for forecasting are commonly used.
With exceptions \citep{gornitz2013toward, gao2020robusttad, carmona2021neural}, AD is typically treated as an \emph{unsupervised} machine learning problem, i.e.\ the training data contains both normal and anomalous instances, but no labels to indicate which is which.
In this paper, we follow this paradigm and address the problem of unsupervised time series AD by using a probabilistic time series model based on  Gaussian processes~(GP). 


Recent research in time series AD has focused on improving the detection performance in the large-data regime by leveraging advances in deep learning~(DL)~\citep{su2019robust, ren2019time, shen2020timeseries, audibert2020usad, carmona2021neural}.
These models can yield high accuracy when sufficient amounts of data are available for training but are not universally applicable. 
%
For example, in embedded systems, hardware capabilities or latency requirements may impose severe limitations on model selection.
 Furthermore, it is often beneficial to encode prior knowledge about the types of \emph{relevant} anomalies~\citep{gornitz2013toward}, a challenging undertaking when data-driven DL models are used. 
Lastly, more often than not is training data scarce which exacerbates the deployment of data-hungry DL approaches.

Many alternative approaches to deep learning for AD exist and often, heuristics such as thresholds based on statistical moments of the data can be satisfactory. 
An alternative class of algorithms focuses on isolating anomalies rather than on constructing models of the \emph{normal} regime~\citep{liu2012isolation, rrcf}.
In addition, any time series forecasting model can readily be used for anomaly detection, by declaring points that deviate sufficiently from the predictions as anomalous~\citep{chakraborty2020building}.
In this paper we propose an unsupervised online anomaly detection method for uni- and multivariate time series data that emphasizes computational and sample efficiency, while being flexible enough to model complex temporal patterns and correlation structures. 
The multivariate variant of our method is based on a factor-analysis variant of multi-output GPs  that admit a state space representation for temporal data, enabling efficient linear-time inference. 
In particular, we make the following contributions:
\begin{enumerate}
	\item we propose to leverage the state space GP~(ssGP) framework \citep{solin2016statespace} to enable linear-time inference in the multivariate time series model Gaussian process factor analysis \citep{yu2009gaussian}; 
	\item we propose to enforce an orthogonality constraint on the factor loading matrix, which aids interpretability by decoupling the latent processes and provides an additional speed-up; 
	\item we propose a simple heuristic for improving the robustness of the Kalman filter inference in the presence of anomalies;
	\item we show that our approach can not just detect anomalies, but also provides explainability by attributing them to specific latent components or the noise process.  
\end{enumerate}

The remainder of this article is structured as follows: Sec.\ \ref{sec:related} contextualizes our proposal in the existing literature, followed by a description of the necessary background in Sec.\ \ref{sec:background}. We present our method in Sec.\ \ref{sec:method}, followed by quantitative and qualitative experiments (Sec.\ \ref{sec:experiments}) and their results (Sec.\ \ref{sec:results}), concluding with a discussion in Sec.\ \ref{sec:discussion}.

\section{Related Work}\label{sec:related}


We first review work on state space models and GPs for time series AD and contrast our proposal with prior art. Then, we detail approaches that focus on adapting GPs to streaming time series and reduce time complexity of GPs. 

Early approaches of anomaly detection with \emph{fixed} state state space models go back to the work by~\citet{chib1994outlier} where anomalies are detected by a threshold on the predicted noise variance and are followed by statistically motivated methods~\citep{soule2005combining}, where the residual of predicted and updated state is treated as a random variable and a statistical hypothesis test is performed to detect anomalies.

GPs have been used in the context of anomaly detection for specific applications such as healthcare monitoring~\cite{chandola2011gaussian, pang2014anomaly}.
%
%
%
The closest work to ours is SGP-Q~\citep{gu2020online}, where a sparse variational GP \cite{titsias2009variational} is used and anomaly detection is performed by maintaining a sliding window of the last $m$ observations.
After fitting a GP on the window, the probability distribution over the next point is obtained.
If this observation exceeds a likelihood threshold under the predictive distribution, it is included in the window, otherwise the mean of the predictive distribution is included.
This method differs from ours in four key aspects:
\begin{inparaenum}
	\item Our model allows for faster inference without the need of inverting a~(potentially large) kernel matrix. 
	\item Deciding on the window size parameter is non-trivial and increases memory requirements, which for our model, are solely determined by the~(typically small) state space dimension.
	\item Our method extends to the multivariate setting and has an inherent explainability component.
	\item While we use a similar robustification procedure, ours allows us to treat unlikely points as missing, naturally leading to an increase in the uncertainty of the predictive distribution.
\end{inparaenum}
These statistical robustification approaches~\citep{huber2004robust} are essential to time series AD and were investigated in the context of Kalman filtering before. 
They range from elaborate multi-step approaches like the ones by \citet{mu2015novel} or \citet{gandhi2009robust} to the approach by \citet{xie1994robust} who derive a stable state estimator with bounded error estimates. 
In the context of anomaly detection \citet{aubet2021monte} propose a principled Bayesian framework to infer which points of the training set are anomalous, however we prefer to have a robustification procedure directly for the Kalman filtering algorithm.

While the effectiveness of GPs was frequently demonstrated on time series tasks, they do not trivially extend to the streaming setting.
Until recently, only little attention had been given to adapting GPs to the streaming setting.
\citet{turner2012gaussian} proposes a method to drastically speed up inference on equally spaced time series.
The linear time complexity sparse variational approximation of GPs~\citep{titsias2009variational, hensman2013gaussian} has been adapted to the setting where data arrives sequentially \cite{bui2017streaming}.
However, the cost of updating the variational parameter would be prohibitive in a streaming setting.
%
%
Speeding up the inference in linear multi-output GPs has been proposed using the sparse variational approximation~\cite{adam2016scalable, duncker2018temporal}, but, to the best of our knowledge, has not been proposed using ssGPs.
Lastly, \citet{hou2019multi} combine the training of conventional GPs with the usage of ssGPs for online camera pose estimation.

\section{Background}\label{sec:background}


%

\subsection{Problem statement}

Let $\vecVar{y}_{:, 1:T} = [\vecVar{y}_{:, 1}, \vecVar{y}_{:, 2}, \ldots, \vecVar{y}_{:, T}]$, denote a multivariate time series, where at each time point $t$%
\footnote{While our method is applicable to non-equally-spaced time series, we abuse notation and conflate time point and time index here to ease the exposition.}%
we have an observation vector $\vecVar{y}_{:, t} = [y_{1, t}, y_{2, t}, \ldots, y_{D, t}]^T \in \mathbb{R}^D$. Our goal is to decide for each $\vecVar{y}_{:, t}$ whether it is anomalous or not, determining the value of the binary \emph{anomaly indicator} variable $a_t \in \{0, 1\}$. In the online AD setting, observations arrive one at a time, and the anomaly decision for point $\vecVar{y}_{:, t}$ can be based only on observations up to and including that point. 

We follow a large body of prior work \citep{ruff2020unifying} and approach this problem via a probabilistic model: for each point $\vecVar{y}_{:, t}$ we compute an anomaly score $s_t = -\log p(\vecVar{y}_{:, t}|\vecVar{y}_{:, 1:t-1})$ based on the predictive density under a model, and determine $a_t$ by thresholding this score, i.e.\ $a_t = \mathbb{I}[s_t > \alpha]$ for some fixed decision threshold $\alpha$. 

\subsection{State Space Gaussian Processes}\label{sec:ssGP}

\label{sec:GP}

\paragraph{A Gaussian process}~(see ~\cite{gp_for_ml} for an exhaustive introduction) is a probability distribution over functions for which any finite number of points have a joint Gaussian distribution.
A GP is fully specified by its mean function $\mu(\xvec): \mathbb{R}^p \rightarrow \mathbb{R}$ and covariance (or kernel) function $k(\xvec, \xvec'): \mathbb{R}^p \times \mathbb{R}^p \rightarrow \mathbb{R}$:
\begin{equation}\label{eq:gp}
	f(\xvec) \sim \mathcal{GP}(\mu(\xvec), k_{\theta}(\xvec, \xvec')).
\end{equation}
In applications of GPs to time series modeling, the input $\xvec$ is typically taken to be the time index $t$, so that the input space dimension is $p=1$, and the observations are modeled as noisy observations of the function values, i.e. $y_t = f(t) + \varepsilon_t$, with $\varepsilon_t \sim \mathcal{N}(0, \sigma^2)$. 
The choice of covariance function and its parameters $\theta$ determine the properties (e.g. smoothness or periodicity) of the resulting random functions, and can be used to encode prior knowledge into the model. 

\label{sec:SSM}

\paragraph{State space models} (SSMs) in general, and linear-Gaussian state space models (LG-SSMs) in particular, are another widely used approach for modeling time series data \citep{durbin2012time, roweis1999unifying}. 
A general LG-SSM is defined as follows:
\begin{align}
	\label{eq:SSM_likelihood}
	\left. \B{y}_{:,t} \middle| \bm{z}_{:,t}\right. &\sim \N \left(\matVar{C} \,\bm{z}_{:,t} + \B{d}, \Psi \right)
	\hspace{30pt}
	\left. \bm{z}_{:,t} \middle| \bm{z}_{:,t-1}\right. \sim \N \left(\matVar{A}\,\bm{z}_{:,t-1}  , \matVar{Q} \right)
\end{align}
where a latent process $\bm{z}_{:, t} \in \mathbb{R}^K$ evolves according to linear dynamics parametrized by the \emph{transition matrix} $\matVar{A}$ driven by Gaussian noise with covariance $\matVar{Q}$. The observations $\bm{y}_{:, t}$ are a linear function of the latent state plus Gaussian noise with covariance $\Psi$. Inference~(e.g.\ computing the distribution $p(\bm{z}_{:, t}|\bm{y}_{:, 1:t})$, known as \emph{filtering}) and likelihood computations in this model can be performed in closed form and in linear time~(in the number of time steps) via the well-known Kalman filter algorithm \citep{kalman1960new}, and parameters can be learned via maximum likelihood, expectation maximization~(EM) \citep{dempster1977maximum}, or via spectral methods~(see e.g.\ \citep{roweis1999unifying, durbin2012time} for details).

\paragraph{State Space Gaussian Processes} 
Linear-Gaussian state space models and GPs are closely connected, and in some cases can be seen as two different views on the same underlying model. 
In particular, for a large class of commonly-used kernel functions, a temporal Gaussian process model can equivalently be expressed in the form of a LG-SSM, where the state transition matrix $\matVar{A}$ and covariance matrix $\matVar{Q}$ are derived from the GP kernel and its parameters.
The main practical benefit of this conversion is that the model can still be conveniently specified in the form of a covariance kernel, while inference can be performed in linear time using Kalman filtering in the state space representation.
More precisely, for a large class of covariance functions $k(t, t')$~(see \citet{solin2016statespace} for details) a temporal GP model of the form
\begin{align}
\label{eq:gp_temporal}
	f(t) & \sim \mathcal{GP}(\mu(t), k_{\theta}(t, t'))
	\hspace{30pt}
	y_t = f(t) + \varepsilon_t \quad \quad \quad \varepsilon_t \sim \mathcal{N}(0, \sigma^2)
\end{align}
can equivalently be expressed as a LG-SSM of the form
	\begin{align}
		\label{eq:discModel}
		\vecVar{f}_\timeVar &= \matVar{A}_{\timeVar-1} \vecVar{f}_{\timeVar-1} + \matVar{q}_{\timeVar-1} \text{, with } \vecVar{q}_{\timeVar-1} \sim\N(\matVar{0}, \matVar{Q}_{\timeVar-1}) \\
		\label{eq:discModel_obseravtion}
		y_\timeVar &= \vecVar{h}^T \vecVar{f}_\timeVar + \epsilon_t \text{, with } \epsilon_t \sim \N(0, \sigma^2)
	\end{align}

%
where the state transiton matrix $\matVar{A}_{\timeVar-1}$, the state noise covariance $\matVar{Q}_\timeVar$, and the emission vector $\vecVar{h}$ are determined by the kernel function $k(t, t')$ and -- in the non-equally-spaced setting -- depend on the time since the last observation was made. 
As this is a standard LG-SSM, Kalman filtering and smoothing can be used to obtain the posterior distribution over the latent variable $\vecVar{f}_t$. 

\subsection{Gaussian Process Factor Analysis}
GPs have been extended to vector-valued functions in the form of multi-output Gaussian processes~(MOGPs)~(see e.g.\ \citep{bruinsma2020scalable} and references therein).
%
To avoid computational complexity scaling cubically both in the number of points $T$ and the number of output dimensions $D$, i.e. $\mathcal{O}(T^3 D^3)$,  one can make the assumption that the data can be explained as a linear combination of a small number of latent \emph{factors}. Such \emph{factor analysis} models have a long history and have been explored in the i.i.d.\ setting \citep{roweis1999unifying}, in the context of state space models \citep{jungbacker2008likelihood}, and in the context of GPs \citep{teh2005semiparametric, yu2009gaussian}.


The resulting model, called Gaussian process factor analysis (GPFA) \citep{teh2005semiparametric, yu2009gaussian}, can be described as follows. 
Characteristically for a factor analysis model, there exists a linear-Gaussian relationship between the $D$-dimensional observations $\vecVar{y}_{:, t}$, and a $K$-dimensional latent state $\vecVar{z}_{:, t}$. Each of the $K$ dimensions of the latent state time series is given an independent GP prior, i.e., 
\begin{align}
	\label{eq:GPFA_likelihood}
	\left. \B{y}_{:,t}\middle| \bm{z}_{:,t}\right. &\sim \N \left( \matVar{C} \,\bm{z}_{:,t} + \B{d}, \Psi \right) \\
	\label{eq:GPFA_prior}
	\bm{z}_{k,:} &\sim \N (0, K_{\B{\tau}\B{\tau}}^{(k)} ) \quad \quad k=1,\ldots, K
\end{align}
where $\matVar{C} \in \mathbb{R}^{D\times K}$ maps the latent to the observed space with an offset $ \bm{d} \in \mathbb{R}^{D\times 1} $ and $\Psi \in \mathbb{R}^{D\times D} $ is a diagonal covariance matrix (as in classical factor analysis, each element of its diagonal is the independent noise variance of the corresponding output dimension). $ \bm{z}_{k,:} \in \mathbb{R}^{1 \times T} $ is the time series corresponding to latent dimension $k$, and $ K_{\B{\tau}\B{\tau}}^{(k)} \in \mathbb{R}^{T \times T} $ is the covariance matrix of the GP prior on the $k^{th}$ latent dimension, and $\B{\tau} = \{1, 2, \dots , T\}$ the set of all the time steps.
Each latent process has a different set of parameters for its kernel function.




\section{Method}\label{sec:method}


We propose to perform online anomaly detection in multivariate time series using a variant of GP factor analysis \citep{yu2009gaussian} combined with the state space GP framework of \citet{solin2016statespace}, resulting in a method we refer to as \emph{state space Gaussian process factor analysis} (\method).
Taking advantage of the ssGP framework allows us to efficiently perform (online) inference and learning in the resulting state space model using Kalman filtering in time linear in the number of time steps. To further reduce computational complexity and to gain the ability to explain the anomalies in terms of latent causes, we constrain the columns of the \emph{factor loading matrix} $\matVar{C}$ to be orthogonal, following ideas proposed by \citep{bruinsma2020scalable} in the context of MOGPs. 
Finally, we propose a heuristic to improve the robustness of the inference procedure specifically for anomaly detection on streams of data, by foregoing the latent state update for outliers. 

%

\subsection{State Space Gaussian Processes Factor Analysis}
Our underlying probabilistic time series models is a combination of the GPFA model and ssGPs.
%
In particular, we replace each of the $K$ independent GPs in eq.~\eqref{eq:GPFA_prior} of the GPFA model with its corresponding ssGP state space model, whose latent state $\vecVar{f}_t^{(k)}$ evolves according to eq.~\eqref{eq:gp_temporal}. The latent factor $\bm{z}_{k,t}$ is then given by $ \bm{z}_{k,t} = \vecVar{h}^{(k)T} \vecVar{f}_t^{(k)} $.


This allows us to rewrite eq.~\eqref{eq:GPFA_likelihood} in terms of the latent state of the corresponding ssGP, 
\begin{align}
	\label{eq:SSGPFA_likelihood}
	\left. \B{y}_{:,t}\middle| \bm{z}_{:,t}\right. &\sim \N \left( \sum_{k=1}^{K} \matVar{C}_{:,k} \, \bm{z}_{k,t} + \B{d}, \Psi \right)
	\hspace{10pt} \sim  \N \left( \sum_{k=1}^{K} \matVar{C}_{:,k} \, \vecVar{h}^{(k)T}  \vecVar{f}_t^{(k)} + \B{d}, \Psi \right)
	\\
	\nonumber
	&\sim  \N \left(  \begin{bmatrix} \matVar{C}_{:,1} \vecVar{h}^{(1)T} &  ... & \matVar{C}_{:,K} \vecVar{h}^{(K)T} \end{bmatrix}  \begin{bmatrix}
		\vecVar{f}_t^{(1)} 
		\\
		\vdots 
		\\
		\vecVar{f}_t^{(K)} 
	\end{bmatrix} + \B{d}, \Psi \right)
\end{align}
where $ \matVar{C}_{:,k}$ is the $k^{th}$ column of $\matVar{C}$. This reformulation allows us to treat the whole model as one linear state space model in which the state is the concatenation of the states of the $K$ latent ssGPs. The transition and state noise covariance matrices are block diagonal matrices containing respective matrices of the corresponding latent ssGP:
\begin{align*}
	\tilde{\matVar{A}}_{t}  &= \begin{bmatrix}
		\matVar{A}_t^{(1)} &  ... & \B{0}
		\\
		\vdots & \ddots & \vdots
		\\
		\B{0} &  ... & \matVar{A}_t^{(K)}
	\end{bmatrix}
\hspace{50pt}
\tilde{\matVar{Q}}_{t}  = \begin{bmatrix}
	\matVar{Q}_t^{(1)} &  ... & \B{0}
	\\
	\vdots & \ddots & \vdots
	\\
	\B{0} &  ... & \matVar{Q}_t^{(K)}
\end{bmatrix}
\end{align*}

This way, we benefit from all the advantages of ssGPs and GPFA: we can model a multi dimensional time series as lower dimensional processes with different kernel functions and still employ efficient Kalman filtering to use the model on streams of data. 
We can use the EM algorithm to fit the model parameters as well as the hyperparameters of each of the $K$ kernel functions.
In the E-step we use the Kalman filtering and smoothing to obtain the posterior on the latent variables and the M-step is equivalent to the original GPFA model.

\subsection{GPFA with Independent Latents}

To allow for explainability through disentanglement and to speed up training and inference, one can force the latent processes to be independent of each other a posteriori.
We show how this can be achieved in the standard GPFA, here for the inference of the latents for a single time point $t$, this extends to the inference to the whole time series as we use the E-step formulation presented by \citet{yu2009gaussian}.
We write the prior on $ \bm{z}_{:,t}$ as:
\begin{align}
	\label{eq:GPFA_prior_time}
	\bm{z}_{:,t} &\sim \N (0, \tilde{K}_{tt} )
\end{align}
where $\tilde{K}_{tt}$ is a diagonal matrix where the $k^{th}$ entry is the prior variance on $ \bm{z}_{k,t}$ given by the kernel function evaluated at this point.
Using Gaussian conditioning with equations \ref{eq:GPFA_likelihood} and \ref{eq:GPFA_prior_time} it follows that the posterior on the latent at time step $t$ is given by:
\begin{align}
	\left. \bm{z}_{:,t}  \middle|  \B{y}_{:,t}  \right. &\sim \N \left( \Sigma_t \, \matVar{C}^T \Psi\inv \, (\B{y}_{:,t}  - \B{d}), \Sigma_t \right) 
	\\
	\Sigma_t &= \left( \tilde{K}_{tt}\inv +  \, \matVar{C}^T \Psi\inv \, \matVar{C} \right)\inv
\end{align}
For the latent to be independent a posteriori, we need $ \Sigma_t $ to be diagonal. As the processes are independent a priori, $ \tilde{K}_{tt} $ is diagonal.
When a matrix has orthogonal columns, then $\matVar{C}^T  \, \matVar{C}  = I$. 
Therefore, the right hand side term is diagonal if one constrains $\matVar{C}$ to have orthogonal columns and  constrains  $\Psi$ to be written as $I \sigma^2 $:
\begin{align}
	\matVar{C}^T \Psi\inv \, \matVar{C} = \matVar{C}^T I \sigma^2 \, \matVar{C} = \matVar{C}^T  \, \matVar{C} I \sigma^2 = I \sigma^2
\end{align}

Constraining the columns of $\matVar{C}$ to be orthogonal means that one cannot simply use closed form update in the M-step~($\matVar{C}^*$).
At each M-step, we propose to set $\matVar{C}$ to the closest orthogonal matrix to $\matVar{C}^* $  in the Frobenius norm.
This is also the closest matrix in the KL divergence between the likelihood of the GPFA and the orthogonalised GPFA, since \cite{bruinsma2020scalable} showed that the distance in this KL is proportional to the Frobenius norm between the two matrices. This means that we obtain the orthogonal $\matVar{C}$ that maximises the free energy at each M-step.

The closest orthogonal $\matVar{C}$ to $\matVar{C}^* $  in the Frobenius norm is obtained with singular value decomposition \citep{higham1988matrix}:  $\matVar{C}^*  = UDV $, where $U \in \R^{N\times K}$ and $V \in \R^{K\times K}$ have orthogonal columns, and $D \in \R^{K\times K}$ is diagonal. We obtain $\matVar{C} = U V^T$ which has orthogonal columns.
This saves us any gradient optimisation to update the model parameters.
This procedure is identical to the proposed post-processing step of the original GPFA, with the difference that we do it at every M-step which allows the learned model to give this intepretability.

This constraint allows us to fit each of the latent processes in the E-step in parallel yielding substantial speed gains. In this case, we have a combination of $K$ state space models for which the state and the state transition matrix are the same as the ones of the corresponding latent GP, we only adjust the emission matrix to map to the observed dimension through the column of $\matVar{C}$.

Note that, while the likelihood of FA is not affected by the orthogonalisation of the columns of the loading matrix~(proof in Sec.~\ref{sec:app FA C}), this is not the case in GPFA, where the likelihood is given by:
$ p(\B{y}_{:,t}| \theta) = \N(\B{y}_{:,t} | \bm{d}, \Psi + \matVar{C} \tilde{K}_{tt} \matVar{C}^T ) $.
We see that orthogonalising the columns of $\matVar{C}$ \emph{does} change the likelihood. In particular, if the unconstrained maximum likelihood loading matrix $\matVar{C}$ does not have orthogonal columns, the orthogonalisation step will lead to a decrease in likelihood.
We argue that in practice, other assumptions tied to the GPFA model, like the linear mapping from the latent to the observed, may also not hold in the true generating process, and so that the additional assumption of having independent latent processes may be acceptable in practice.

\subsection{Interpretability and Explainability of Detected Anomalies}


\method\ allows us to detect anomalies in individual dimensions of a multivariate time series~(through the marginal on each dimension) as well as in the whole sequence~(through the joint distribution).
When constraining the latents to be orthogonal, we can attribute the contribution of each latent dimension to the final anomaly score. In doing so, we can determine ``the origin'' of an anomaly and provide an interpretation if the latent processes can themselves be interpreted~(e.g.\ the period parameter in Periodic kernels).

Given an anomaly occurring at time $t$, we rely on the orthogonal columns of $\matVar{C}$ to obtain the latent values $\B{v}_{:,t} $ that would have best explained the observed point.
These values are the ones that minimise $ \B{y}_{:,t} - \matVar{C} \B{v}_{:,t}  $ i.e. $\B{v}_{:,t} = \left( \matVar{C} \matVar{C}^T\right)\inv \matVar{C}^T \B{y}_{:,t}  $ . The least squared solution is unique when the columns of $\matVar{C}$ are orthogonal as it results in the dimensions of $ \B{v}_{:,t}  $ to be independent of each other.

We can compute the likelihood of $ \B{v}_{k,t}  $ under the probability distribution on $ \bm{z}_{k,t}  $, the corresponding latent process distribution. This likelihood quantifies the contribution of the latent to a point having been classified as anomalous. 
%
%
It may also happen that detected anomaly cannot be explained by the sub-space of the observed space spanned by the columns of $\matVar{C}$. In this case, the reconstruction error $ \B{y}_{:,t} - \matVar{C} \B{v}_{:,t}  $ will be high. This indicates that we cannot rely on our latent processes to explain the cause of the anomaly~(see Sec. \ref{subsec:explainable} for an illustration).



\begin{algorithm}[tpb]
	\SetAlgoLined
	\DontPrintSemicolon
	\KwIn{Time series $\B{t}=( \B{y}_\timeVar)_{\timeVar=1}^{T}$, update threshold $\rho$
	}
	$o:= 1$ \tcp*[f]{\small Index last relevant observation} 
	
	\For{$t \in \{1, \dots, T\}$}{
			$\Delta_t \gets t - o$
		
		$\matVar{A}_t = \exp{(\matVar{F}\Delta_t)}$\\
		$\matVar{Q}_t = \mathbf{P}_\infty-\mathbf{A}_t\mathbf{P}_\infty\mathbf{A}_t^T$
		
		\tcp{Kalman prediction}
		
		$\matVar{m}_{t|t-1}$, $\matVar{P}_{t|t-1} \gets$ prediction with $\text{Eq.}~\ref{eq:discModel}$
		
		
		$  \log\{p(\B{y}_t)\}  \gets $ likelihood of observed with $\text{Eq.}~\ref{eq:discModel_obseravtion}$

		
		
		\tcp{Robustify}
		\lIf{$\log{p(\B{y}_t)} > \rho$}{
			
			
			\hspace{10pt}	$\matVar{m}_\text{t|t}$, $\matVar{P}_\text{t|t}\gets$ filtering mean and covariance \\			
			\hspace{10pt}	$o \gets t$  \tcp*[f]{\small set o to last normal index}
		}
	}
	\caption{Robust Kalman filtering}
	\label{alg:rKF}
\end{algorithm}

\subsection{Specificities to Online Anomaly Detection}

%
%
%
%

We define the anomaly score of an observation as its negative log-likelihood under the predictive distribution generated for that point which makes each score interpretable.

\paragraph{Robust Kalman Filter}

State space GPs are commonly used for regression or forecasting tasks where one can assume that all observations are normal.
In the AD setting, some points may also be anomalous.
We propose a simple and intuitive heuristic to address this problem. If a point is too unlikely, it is not used for the state update, and therefore treated as missing.
We introduce a user-defined update threshold $\rho$, controlling how likely, given the current model, an observation must be to contribute to the update~(see lines 7 \& 8 of Algorithm~\ref{alg:rKF}).
%
%
The advantages of this method over the ones introduced in Sec.~\ref{sec:related} lie in its simplicity and that it leads to an increase in the predictive uncertainty after ignored points.

\paragraph{Computational and memory complexity}
The computational complexity for \textbf{training} \emph{with} orthogonalisation constraint on $\matVar{C}$ is $\mathcal{O}(T D^3 K L^2 + T K L^3)$, and $\mathcal{O}(T D^3 K^2 L^2 + T K^3 L^3)$ without constraints. $L$ is the maximum state space dimension among all latent processes.
For \textbf{Inference}, the complexity is $\mathcal{O}( D^3 K L^2 + K L^3)$ if $\matVar{C}$ is constrained, and $\mathcal{O}(D^3 K^2 L^2 + K^3 L^3)$ otherwise ~(see section \ref{sec app:complexity} of the appendix for more details).
Beyond making the complexity scale linearly in the number of latent dimensions, the orthogonalisation of $C$ allows to compute the filtering, smoothing, and the E-step for each latent process independently in parallel.

\section{Experiments}\label{sec:experiments}


\subsection{Explainable Anomaly Detection}
\label{subsec:explainable}

\begin{figure*}[!t]
	\begin{subfigure}[b]{.5\textwidth}
		\includegraphics[height=3.5cm]{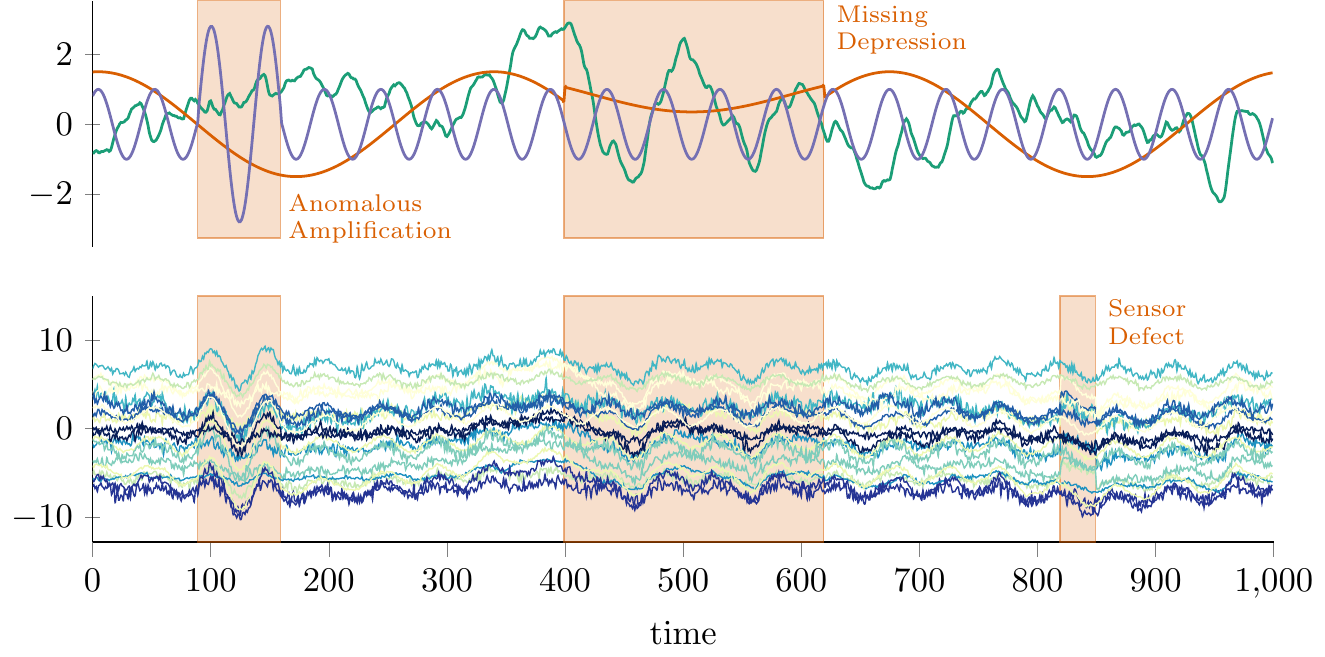}  
		\caption{Generating latent processes \& observed time series}
		\label{fig:sub-first}
	\end{subfigure}
	\begin{subfigure}[b]{.5\textwidth}
		\includegraphics[height=3.5cm]{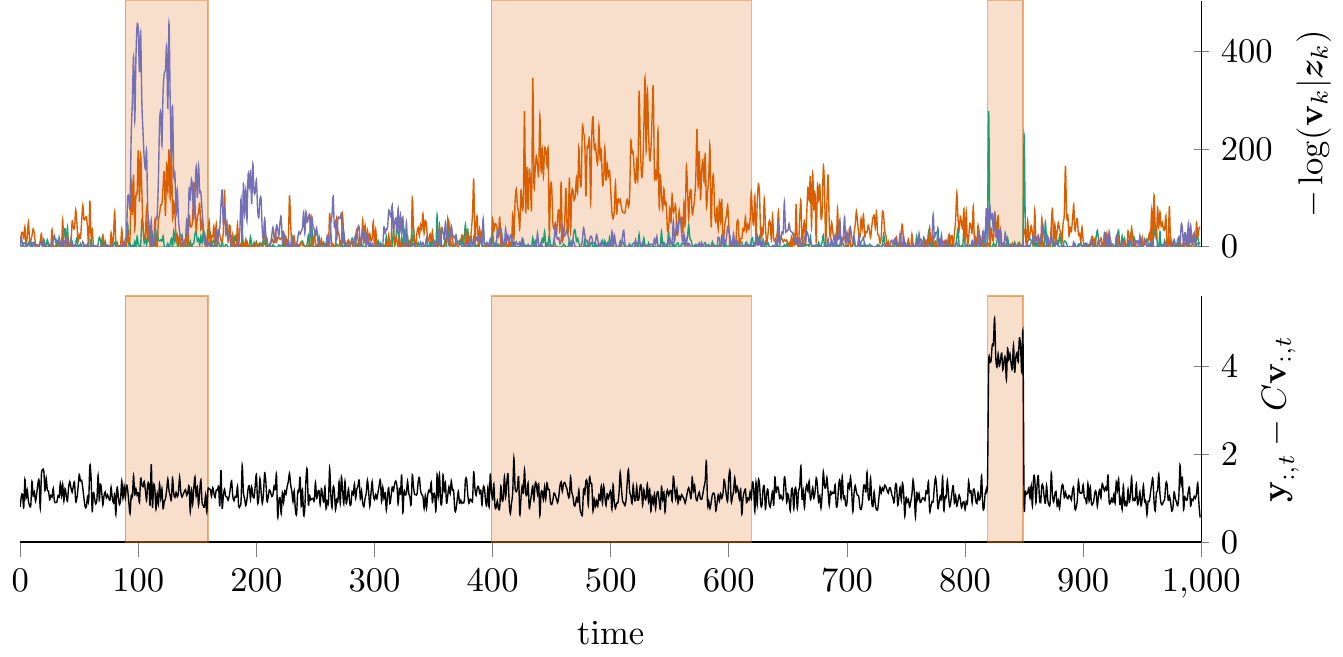}  
		\caption{Neg. log-likelihoods per latent ssGP \& proj. error}
		\label{fig:sub-second}
	\end{subfigure}
	\caption{\small Left: three generating latent processes~(top) and observed time series~(bottom) with anomalous regions.
		Right: negative log likelihood of the projected latent for each dimension~(top) and projection error~(bottom).
	}\label{fig:explainability}
\end{figure*}

We illustrate the explainability capabilities of our model on a synthetic dataset.
For this, we first draw three latent time series: one from a Mat\'ern kernel and two from Cosine kernels whose varying periods simulate daily and weekly fluctuations.
%
%
Anomalies in the \emph{data generating process} are injected by increasing and dampening the short and long period Cosine latents, respectively. 
Furthermore, a measurement offset is added to 7 of the \emph{observed} dimensions simulating a defect sensor.
%
%
%
We train \method\ on time series generated from the same underlying yet uncorrupted process, and Figure~\ref{fig:explainability} depicts the results of running the streaming anomaly detection on corrupted \emph{test} sequences.
The generating latent processes and observed time series are shown at the top and bottom of Figure~\ref{fig:sub-first}.

Figure~\ref{fig:sub-second} depicts how the likelihoods of projected latents $\B{v}_k$ under the predictive distributions of the model's latent process~(top-right) allows to accurately identify the latents that caused the observed point to be classified as anomalous:
The projection of the short period latent has a low likelihood in the first anomaly; the projection of the long period latent process is unlikely in the second anomalous region.
The third anomaly is an irregularity in the coordination of the observed dimensions with each other.
As a result, the observed points are not part of the sub-space that is spanned by the columns of $\matVar{C}$ in the observed space, which causes a high projection error~(bottom-right).

\subsection{Robustness to anomalies}

\begin{figure}[!t]
	\centering
	\includegraphics[width=0.85\linewidth]{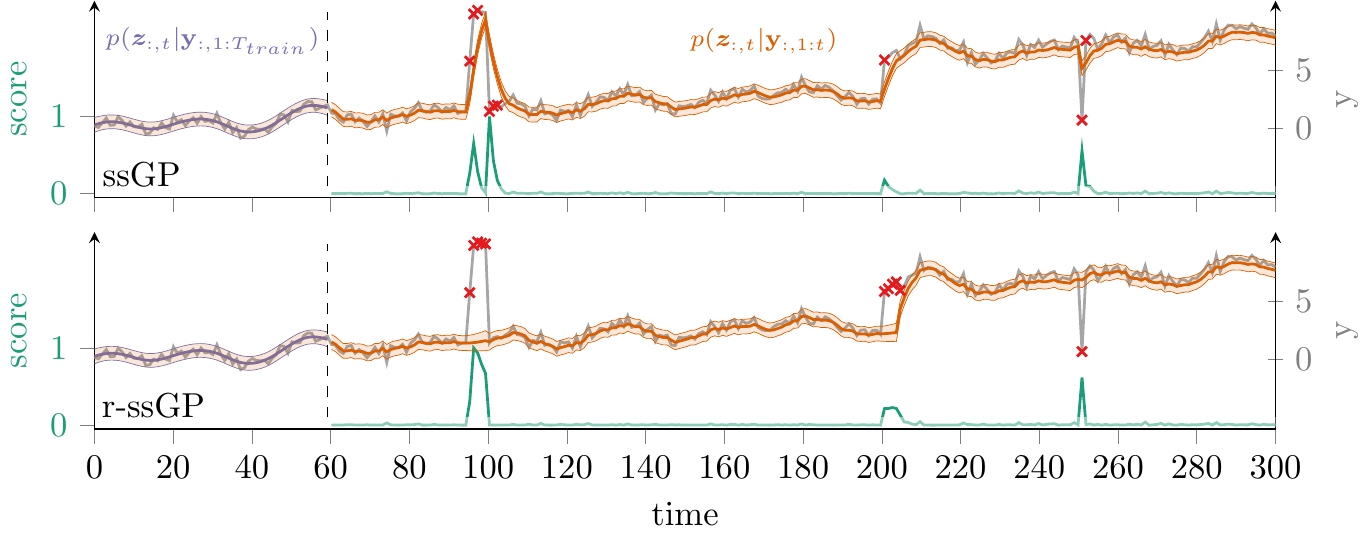}
	\caption{\small Robust~(\methodUniRob) and non-robust~(\methodUni) models on a synthetic time series. Filter distributions are shown in orange, scaled negative log-likelihoods in green. Scores below $0.1$ are marked with red crosses. }
	\label{fig:synth1}
\end{figure}

A demonstration of the effect of the proposed robustification of the Kalman update~(at $\rho=\num{1e-12}$) is shown in Figure~\ref{fig:synth1}.
We generate a synthetic time series with two anomalies and a change point following the equation in ~\ref{sec:app synth data}.
%
%
%
%
Both robust and non-robust ssGPs are trained on the first $60$ data points and their filtering solutions are visualised on the rest of the time series.

%
%
%
The non-robust ssGP rapidly adjusts to anomalies persisting over a longer period~(around $t=100$), resulting in distorted predictions and anomaly scores whereas \methodUniRob\ predictions are unaffected.
%
%
%
When a sustained distribution shift occurs~(around $t=200$), the increasing uncertainty of future predictions allows \methodUniRob\ to adjust once the shift can be considered as a new normal regime.


\subsection{Detection Performance}
%
\paragraph{Evaluating Anomaly Detectors}

Label sparsity prevents employing performance metrics such as accuracy.
Commonly used metrics are the $F_1$-score, precision, and recall redefined for anomaly ranges:
Consecutive anomalies are bundled into \emph{one} anomaly range.
If a single observation in this range is detected, all observations in this range are considered to be correctly identified.
%
%
For each method, we perform a search for the best $F_1$-score by considering all unique scores a method generated for the data set at hand as decision thresholds.
%
%
We are well aware of the valid criticism of this evaluation method by \citet{kim2021towards}, however we propose here to demonstrate that our method performs reasonably rather than set a new state-of-the-art result. We think that for our purposes this metric is sufficient.

\paragraph{Data Sets \& Preprocessing}
We use three publicly available data sets for time series anomaly detection to assess detection performance.
%
%
The multivariate data sets include the NASA data sets~(SMAP and MSL)~\citep{nasa} and the Server-Machine Dataset~(SMD)~\citep{su2019robust}.
The Numenta Anomaly Benchmark (NAB)~\citep{nab} is a univariate data set on which we evaluate \methodUni, and \methodUniRob.
%
%
SMAP and MSL contain telemetry data from two NASA spacecrafts and NAB, among others, network utilisation and server temperature measurements.
%
SMAP, MSL and SMD are already partitioned in an unlabeled train set and a test set,
whereas, NAB has no predefined split.
We therefore split each time series into a \SI{20}{\percent} training and \SI{80}{\percent} testing sequence.
%
%
%
For more information about the data sets, we refer to Table~\ref{tab:app stats} in the appendix.
We standardise the training sequence to have zero mean and unit variance and use its mean and standard deviation to scale the test sequence accordingly. 
 
%
%

\paragraph{Comparison Partners}

On NAB, we compare to \rrcf~\citep{rrcf} and \htm. The former is a random cut tree based method commonly used by practitioners and the later achieves state-of-the-art performance on univariate time series. The anomaly scores of \htm\ are taken directly from their publication to compute its performance.
For SMD and NASA, we compare to \omni~\citep{su2019robust}, NCAD~\citep{carmona2021neural}, USAD~\citep{audibert2020usad}, THOC~\citep{shen2020timeseries}, LSTM-NDT~\citep{nasa}, DAGMM~\citep{dagmm}, and LSTM-VAE~\citep{lstmvae}.
The first four methods represent state-of-the-art neural network approaches.
\omni\ combines different ideas from RNNs, variational autoencoders, and normalizing flows, while THOC combines spherical embeddings on multi-scale temporal features with one-class classification. NCAD on the other hand allows to incorporate labels into the one-class classification approach. We take all performance scores from \citet{su2019robust} and the respective papers of the methods, we run \rrcf~ ourselves.

\paragraph{Implementation and Hyperparameters}
We implemented our methods using PyTorch~\citep{pytorch} and GPy's~\cite{gpy2014} SDE kernels as a reference implementation.
For all multivariate experiments, we used four versatile Mat\'ern kernels~($\nu = \frac{3}{2}$) with fixed parameters~(see Section~\ref{sec:app impl details} for details)
For the univariate data set we combined an integrated Brownian motion kernel to cover random fluctuations, a Cosine Kernel to cover periodic behaviour, and a versatile Mat\'ern~($\nu = \frac{3}{2}$).
An example of the state space representation of the Mat\'ern kernel is shown in Section~\ref{sec:app matrices}, more can be found in~\cite{solin2016statespace}.
%

\section{Results}\label{sec:results}

\begin{table*}[!t]
	\caption{\small Detection performance on the SMD, MSL, and SMAP benchmark data sets.}
	\label{tab:smd results}
	\centering
	\sisetup{%
		mode         = text,
		table-format = 2.2,
		table-figures-decimal = 2
	}
	\scriptsize
	\begin{tabular}{%
			lSSS
			SSS
			SSS
			S%
			S[separate-uncertainty=true,
			open-bracket = ,
			close-bracket = ,]
		} \toprule
		& \multicolumn{3}{c}{SMD} & \multicolumn{3}{c}{MSL} & \multicolumn{3}{c}{SMAP} & & {Inference time} \\
		& {$F_1$} & {Prec.} & {Rec.} & {$F_1$} & {Prec.} & {Rec.} & {$F_1$} & {Prec.} & {Rec.} & {\# Params} & {per time point} \\ \midrule
		{\rrcf} & 55.17 & 50.27 & 61.13 & 91.35 & 88.83 & 94.01 & 92.20 & 92.09 & 92.31 & {0}  &  \SI{8.17 \pm 0.86}{\milli\second} \\ 
		{OmniAnomaly} & 88.57 & 83.34 & 94.49 & 89.89 & 88.67 & 91.17 & 84.34 & 74.16  & 97.76 & {$\approx 2.6$ Mio.} & \SI{1.55 \pm 0.0}{\milli\second}  \\ 
		{NCAD} & 80.16 & 76.08& {--} & 95.47 & 94.81 & 96.16 & 94.45 & 96.24 & 92.73 & {$\approx 32.6$ k} & {--}\\
		{USAD} & 93.82 & 93.14 & 96.17 & 91.09 & 88.10 & 97.86 & 81.86 & 76.97 & 98.31 & {--} & {--}\\
		{THOC} & {--} & {--} & {--} & 93.67 & {--} & {--} & 95.18 & {--}  & {--} & {--} & {--}\\
		{LSTM-NDT} & 60.37 & 56.84 & 64.38 & 56.40 & 59.34 & 53.74 & 89.05 & 89.65  & 88.46& {$\approx 100$ k} & {--}\\
		{DAGMM} & 70.94 & 59.51 & 87.82 & 70.07& 54.12  & 99.34 & 71.05 & 58.45 & 90.58 & {--} & {--} \\
		{LSTM-VAE} & 78.42  & 79.22  & 70.75 & 67.80  & 52.57  & 95.46 & 72.98 & 85.51 & 63.66 & {--} & {--} \\ \midrule
		\methodRob~(ours) & 73.15 & 67.35 & 80.05 & 73.94 & 60.93& 94.03 & 71.20 & 97.15 & 56.19 & {236} & \SI{0.78 \pm  0.0}{\milli\second}\\ 
		\method~(ours) & 66.93 & 56.50 & 82.07 & 66.05 & 55.64 & 81.26 & 66.54 & 81.15 & 56.39 & {236} & \SI{0.78 \pm 0.0}{\milli\second} \\ \bottomrule 
	\end{tabular}
\end{table*}

Table~\ref{tab:smd results} shows the performance of our model on the \emph{multivariate} data sets.
We note the strong positive effect of the robust filtering update on the precision, and overall $F_1$-score.
This is a particularly desirable property for applications where false positives are very costly. 
Overall, \methodRob's performance is competitive with its deep contenders. 
Interestingly, despite its simplicity, \rrcf, a method largely ignored in recent AD works, reaches high performance values on the NASA data sets.
%
%
The fact that our method with its low number of parameters outperforms a deep method such as DAGMM in all three data sets underlines the effectiveness of our approach.
Lastly, our method exhibits the fastest inference time, outperforming \rrcf\ by an order of magnitude.


\begin{table}[!h]
	\caption{\small Detection performance on the univariate NAB benchmark data set.}\label{tab:nab results}
	\centering
	\sisetup{%
		mode         = text,
		table-format = 2.2,
		table-figures-decimal = 2
	}
	\scriptsize
	\begin{tabular}{lSSSSS[table-format=2]} \toprule
		& {$F_1$} & {Prec.} & {Rec.} & {\# Params} & {\# Hyperparams} \\ \midrule
		RRCF & 82.93 & 76.21 & 90.94 & {0} & 3 \\
		HTM & 92.12 & 95.98 & 88.57 & {$64$k} & 17 \\
		\methodUniRob~(ours) & 87.61 & 85.65 & 89.66 & {6} & 1 \\
		\methodUni~(ours) & 90.18 & 90.47 & 89.89 & {6} & 0 \\ \bottomrule
	\end{tabular}
\end{table}

Table~\ref{tab:nab results} shows the performance assessment for the \emph{univariate} NAB dataset. 
While \methodUni\ does not beat the state of the art, it outperforms \rrcf, the low-parameter comparison partner by a large margin.
%
%
We observe a performance drop in the robust version of our method, which we attribute to a property of the NAB data set pointed out by~\citet{deepAnT}: Point-anomalies are centred in an ``anomaly window'' of observations that are labelled as anomalous while being normal.
The range-wise $F_1$-score does not take into account as to whether \methodUni\ gets distorted by an anomaly since \emph{normal} observations are labelled abnormal under this anomaly window.
Lastly, we compared to SGP-Q~\citep{gu2020online}, a conceptually similar method, on all time series for which the authors provide $F_1$-scores:
Our method outperforms SGP-Q on 7/9 time series with a median $F_1$ gain of $3.60$.

\section{Discussion \& Limitations}\label{sec:discussion}

We present \method, a method for interpretable and explainable multivariate time series anomaly detection.
Combining the advantages of both GPFA and ssGPs, our method exhibits excellent time and space complexity while maintaining competitive detection performance which makes it an attractive option for AD on edge devices.
As such, it is particularly attractive for applications in developing countries where access to large compute clusters may be limited.
The GP backbone also allows to generate text descriptions to explain detected anomalies in a similar way as done by the automated statistician~\citep{lloyd2014automatic} for time series characteristics.
As with all GP based approaches, however, kernel selection is one of the main disadvantages when prior knowledge about the data is limited.
%
Furthermore, a study of the influence of parameter initialisation and performance consistency wtih resepct to the robustness parameter $\rho$ could shed more light on the inner workings of our method.
We aimed to provide a fast and competitive alternative for AD and thus did not investigate this yet.
%

\clearpage
\newpage

\bibliography{main}


\clearpage
\appendix

\section{GPFA}

\subsection{Unidentifiability of $C$ in Factor Analysis}
\label{sec:app FA C}

It is known that Factor Analysis is not identifiable: multiple solutions provide the same optimal likelihood.
Let $Q$ to be an arbitrary orthogonal rotation matrix of size $K \times K$ satisfying $Q Q^T = I$. We can obtain a different loading matrix $\ddot{C} = C Q$ for which the likelihood is given by:
\begin{align}
	p(\B{y}_{:,t}| \theta) &= \N(\B{y}_{:,t} | \bm{d}, \Psi + \ddot{C} \ddot{C}^T )
	\\
	&= \N(\B{y}_{:,t} | \bm{d}, \Psi + C Q Q^T C^T )
	\\
	&= \N(\B{y}_{:,t} | \bm{d}, \Psi + C C^T )
\end{align}

\subsection{Closed form model parameter updates}
\label{sec:app M-step GPFA}

Using $q(\bm{z}_{:,t}) = \N\left( \bm{z}_{:,t}   |  \bm{\mu}_t , \Sigma_t \right)$, the posterior distribution on the latent at each time step $t$ infered in the E-step, we obtain the closed form updates for the model parameters:
\begin{align*}
	C^*   &=  \left( \sum_{t=1}^{T}  (\B{y}_{:,t} - \B{d} ) \bm{\mu}_t^T  \right)
	\left( \sum_{t=1}^{T} \left( \Sigma_t + \bm{\mu}_t \bm{\mu}_t^T \right)  \right)^{-1}
	\\
	\B{d}^* &= \frac{1}{T} \sum_{t=1}^{T} ( \B{y}_t - C  \bm{\mu}_t     )
	\\
	\Psi &= \frac{1}{T} \sum_{t=1}^{T}  ( (\B{y}_t - C \bm{\mu}_t -\bm{d}) (\B{y}_t - C \bm{\mu}_t -\bm{d})^T  +    C \Sigma_t C^T  )
\end{align*}
These are the model parameters that maximise the evidance lower bound in the M-step.
Note that the derivations and updates are the same as for Factor Analysis.

\section{Computational and memory complexity}
\label{sec app:complexity}

GPs are known to scale very poorly to big datasets but the combination of introduced methods allows for lightweight training and inference.
%
Recall that $D$ is the number of observed dimensions, $K$ the number of latent processes, and $T$ the number of training time steps. We denote $L$ as the maximum state space dimension among all latent processes.


The computational complexity for \textbf{training} is $\mathcal{O}(T D^3 K L^2 + T K L^3)$ when we constrain the columns of $\matVar{C}$ to be orthogonal and $\mathcal{O}(T D^3 K^2 L^2 + T K^3 L^3)$ without constraints.
In both cases it is linear in the number of training points.
When dealing with very high dimensional time series and few latent processes, one can map the input to a $K$ dimensional space proposed by \cite{bruinsma2020scalable} to allow the training complexity to scale linearly in $D$ but cubically in $K$.

\textbf{Inference} requires the computation of the likelihood of the observation and the Kalman update. The computations complexity is
$\mathcal{O}( D^3 K L^2 + K L^3)$ if $\matVar{C}$ is constained as orthogonal otherwise $\mathcal{O}(D^3 K^2 L^2 + K^3 L^3)$ without constraints.
Beyond making the complexity scale linearly in the number of latent dimensions, the orthogonalisation of $C$ allows to compute the filtering and smoothing (and the E-step) for each of the latent process independently of each other. In such setting, it can be easily parallelized only with only little sequential calculation.

\section{Experiments}\label{sec:appendix}

\subsection{State-Space Matrices}\label{sec:app matrices}
In this section, we exemplify the state space representation of the Mat\'ern Kernel with $\nu=\frac{3}{2}$. For a more comprehensive enumeration of other covariance functions, we refer the interested reader to~\citet{solin2016statespace}.
In the following example, $\matVar{F}$ is the feedback matrix of the corresponding SDE, $\matVar{h}^T$ its measurement model, $\matVar{P}_{\infty}$ the stationary state covariance matrix~(i.e.~the state the process stabilises to in infinity), and $\matVar{A}$  the \emph{discrete-time} state transition matrix.


State-space matrices for Matern32-kernel with length-scale $l$, variance $\sigma^2$ and let $\lambda=\frac{\sqrt{3}}{l}$:
\begin{align*}
	\matVar{F} = \begin{bmatrix}
		0 & 1 \\
		-\lambda^2 & -2\lambda
	\end{bmatrix},\ %
	\matVar{h}^T = \begin{bmatrix}
		1 & 0
	\end{bmatrix},\ %
	\matVar{P}_\infty  = \begin{bmatrix}
		\sigma^2 & 0 \\
		0 & \lambda^2 \sigma^2
	\end{bmatrix}, \\
	\matVar{A} = \exp{(\matVar{F}\Delta_t)}
\end{align*}

Furthermore, it was shown~\cite{solin2016statespace, saerkkae_2019} that the addition and multiplication of covariance functions in state space can be expressed in terms of block diagonals and Kronecker products~($\otimes$) and sums~($\oplus$) as shown below.

\paragraph{Kernel Addition} The addition of two kernels $K_1$ and $K_2$ in the state space formulation is defined by stacking their respective state-space matrices:

\begin{align*}
	\matVar{F} = \text{blkdiag}(\matVar{F}_1, \matVar{F}_2)&, \quad\ %
	\matVar{H} = \text{blkdiag}(\matVar{H}_1, \matVar{H}_2),\ \\
	\matVar{P}_\infty  = \text{blkdiag}(\matVar{P_\infty}_1, \matVar{P_\infty}_2)&, \quad \ %
	\matVar{A} = \exp{(\matVar{F}\Delta_t)}
\end{align*}

\paragraph{Kernel Multiplication} The multiplication of two kernels $K_1$ and $K_2$ in the state space formulation is defined by the Kronecker sum of the respective feedback matrices and the Kronecker product of the remaining state matrices:

\begin{align*}
	\matVar{F} = \matVar{F}_1 \oplus \matVar{F}_2&, \quad \matVar{H} = \matVar{H}_1 \otimes \matVar{H}_2,\ \\ 	
	\matVar{P}_\infty  = \matVar{P_\infty}_1 \otimes \matVar{P_\infty}_2 &, \quad\ %
	\matVar{A} = \exp{(\matVar{F}\Delta_t)}
\end{align*}

\subsection{Implementation Details}\label{sec:app impl details}

\paragraph{RRCF}
The \rrcf\ algorithm represents a predefined number of observations~(i.e. shingle size) as a higher-dimensional point.
In a streaming scenario each new observation gives rise to a new shingle which is used to update the random cut forest.
Using reservoir sampling, the tree can be updated in sublinear time which makes this approach particularly attractive for the online setting.
The authors define the collusive displacement score~(\textsc{CoDisp}), a measure of change in model complexity when a new observation is inserted, as their anomaly score.
We run \rrcf\ with $40$ trees of size  $256$ and a shingle size of $4$.

As for the selected kernel for \methodUni, we do not perform a kernel search, but instead choose a versatile Kernel consisting of the addition of a Brownian motion-kernel with a multiplication of a Matern32 and a Cosine-kernel.
For \method, we use four independent ssGPs with fixed Matern32-kernels with lengthscales~(130, 200, 50, 10).
That being said, our method can readily be used for sophisticated Kernel search algorithms such as~\citep{lloyd2014automatic}.

For all methods, we fix the robustification threshold $\rho$ to be $1\mathrm{e}{-12}$ without performaing any parameter search.
We use the L-BFGS-B~\cite{lbfgs} algorithm to fit our models for maximally $20$ epochs.

\subsection{Synthetic Data}\label{sec:app synth data}

The data generating process yielding the time series of Figure~\ref{fig:synth1} is defined by the following equation.

\begin{equation}\label{eq:synth data}
	f(t)=\cos{(0.04t + 0.33 \pi)} \cdot \sin{(0.2t)} +
	\epsilon_{\text{synth.}}
	+ 5/300t,
\end{equation}
with  $\epsilon_{\text{synth.}} \sim \mathcal{N}(0, 0.15)$.

\section{Data Sets}\label{sec:app data sets}
Table~\ref{tab:app stats} shows data statistics on the used benchmark data sets. $\bar{n}$ refers to the mean time series length. 
\begin{table*}[ht]
	\centering
	\caption{Statistics for the used benchmark data sets. The fraction of anomalies refers to point anomalies, not to anomaly ranges.}\label{tab:app stats}
	\begin{tabular}{llrcrl}
		\toprule
		Data Set &          Sub Data Set &  Num. TS & $D$ &  $\bar{n}$ (train/)test& \% Anomalies \\
		\midrule
		NAB &     realAWSCloudwatch &       17 &  1 &   3984 &        9.3\% \\
		 &        realAdExchange &        6 &  1 &   1601 &       10.0\% \\
		 &        realKnownCause &        7 &  1 &   9937 &        9.6\% \\
		 &           realTraffic &        7 &  1 &   2237 &       10.0\% \\
		 &            realTweets &       10 & 1 &   15863 &        9.9\% \\
		 & artificialWithAnomaly &        6 &  1 &   4032 &       10.0\% \\ \midrule
		NASA &         SMAP &   54 & 25 & 2555/8070 &       \SI{12.83}{\percent} \\
		 &          MSL &   27 & 55 & 1785/2730 &       \SI{10.53}{\percent} \\ \midrule
		 SMD &     machine-1 &  8 &   38 & 24296/24296 &       \SI{6.04}{\percent}   \\
		  &     machine-2 &  9 &   38 & 23691/24813 &       \SI{4.34}{\percent}  \\
		  &     machine-3 &  11 &   38 & 26424/26429 &   \SI{2.75}{\percent}  \\
		\bottomrule
	\end{tabular}
\end{table*}

\section{Contributions}

C.B.\ proposed using ssGPs for time series AD and implemented \methodUni\ and \methodUniRob. Together with L.C.\ and F.X.A., he developed the robustification approach. He ran all benchmarking experiments and together with F.X.A.\ created all figures. He was one of the primary writers of the manuscript.
F.X.A.\ proposed and derived \method. He implemented \method\ together with C.B. He proposed to constrain the model to have independent latent processes a posteriori. Together with J.G., he proposed the optimal closed form update of $\matVar{C}$. He created the method allowing to gain explainability of the detected anomalies and implemented it. He was one of the primary writers of the manuscript.

\end{document}